\documentclass{article}
\usepackage{ijcai24}

\usepackage{times}
\usepackage{soul}
\usepackage{url}
\usepackage[hidelinks]{hyperref}
\usepackage[utf8]{inputenc}
\usepackage[small]{caption}
\usepackage{graphicx}
\usepackage{amsmath}
\usepackage{amsthm}
\usepackage{booktabs}
\usepackage{algorithm}
\usepackage{algorithmic}
\usepackage[switch]{lineno}
\usepackage{amssymb}

\usepackage{booktabs}
\usepackage{multirow}
\usepackage{subfig}
\usepackage{array}

\title{MuseCL: Predicting Urban Socioeconomic Indicators via Multi-Semantic Contrastive Learning}

\author{
Xixian Yong
\and
Xiao Zhou\textsuperscript{\rm }\thanks{Corresponding author.}\\
\affiliations
Gaoling School of Artificial Intelligence, Renmin University of China\\
\emails
\{xixianyong, xiaozhou\}@ruc.edu.cn
}

\begin{document}

\maketitle

\begin{abstract}
Predicting socioeconomic indicators within urban regions is crucial for fostering inclusivity, resilience, and sustainability in cities and human settlements. While pioneering studies have attempted to leverage multi-modal data for socioeconomic prediction, jointly exploring their underlying semantics remains a significant challenge. To address the gap, this paper introduces a \underline{Mu}lti-\underline{Se}mantic \underline{C}ontrastive \underline{L}earning (MuseCL) framework for fine-grained urban region profiling and socioeconomic prediction. Within this framework, we initiate the process by constructing contrastive sample pairs for street view and remote sensing images, capitalizing on the similarities in human mobility and Point of Interest (POI) distribution to derive semantic features from the visual modality. Additionally, we extract semantic insights from POI texts embedded within these regions, employing a pre-trained text encoder. To merge the acquired visual and textual features, we devise an innovative cross-modality-based attentional fusion module, which leverages a contrastive mechanism for integration. Experimental results across multiple cities and indicators consistently highlight the superiority of MuseCL, demonstrating an average improvement of 10\% in $R^2$ compared to various competitive baseline models. The code of this work is publicly available at \url{https://github.com/XixianYong/MuseCL}.
\end{abstract}

\section{Introduction}
Urbanization is intricately connected to critical facets of the United Nations Sustainable Development Goals (UNSDGs), affecting energy, environment, economy, climate, etc. \cite{sachs2022sustainable}. By 2020, over 55\% of the global population resided in urban areas, and this trend is projected to persist and intensify in the forthcoming decades \cite{habitat2022world}. Embracing urbanization yields numerous advantages, including a thriving cultural milieu, enhanced job prospects, and improved transportation networks, etc. However, it also begets a host of predicaments and hurdles, such as air pollution, traffic congestion, and escalated energy consumption \cite{zheng2014urban}. To address these challenges and achieve SDGs, gaining a comprehensive understanding of the urbanization phenomenon through fine-grained region profiling and accurate socioeconomic indicators becomes crucial.

\begin{figure}[t]
\centering
\includegraphics[width=1\columnwidth]{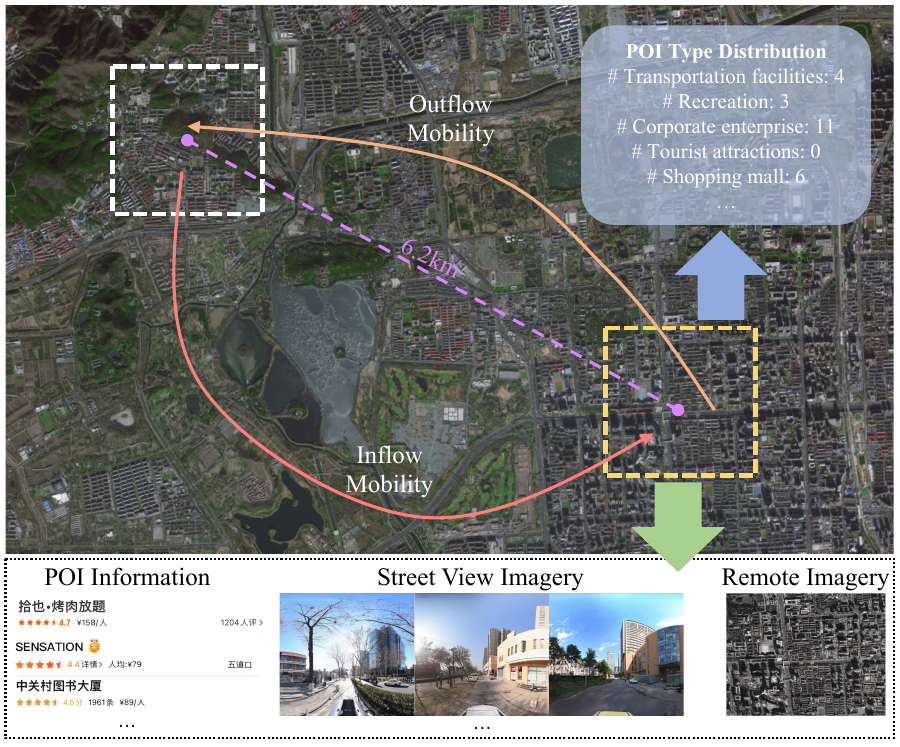}
\caption{Multi-modal data representing urban regions. Each region is linked to its remote sensing and street view imagery, POI data, and inter-region connections, encompassing factors like population mobility, fostering comprehensive insights into the urban landscape.}
\label{data}
\end{figure}

Traditional approaches have relied on community surveys to gather statistics on metrics like population density and household income, which is both resource-intensive and time-consuming \cite{custodio2023review}. With the maturation of urban perception technology, diverse forms of data continue to proliferate within cities, which paves the way for fresh opportunities in tracking urban sustainable development indicators. As depicted in Figure \ref{data}, these datasets encompass point of interest (POI) information, vehicle movement trajectories, remote sensing data, street view imagery, and social media insights, among others. The utilization of this diverse data pool offers robust support for a myriad of downstream tasks. For instance, social media data is instrumental in predicting crime and unemployment rates \cite{antenucci2014using,aghababaei2016mining}, urban lifestyle mining \cite{zhou2018discovering}, and significantly contributes to studies on urban sustainability \cite{ilieva2018social}. Trajectory data reveals valuable insights into mobility patterns, socioeconomic indicators, and health trends \cite{cohen2016using,gao2017identifying,zhou2017cultural,zhou2023phase,wang2018predicting}. POI data aids in discovering new venues to explore \cite{zhou2019topic}, deducing regional functions \cite{yuan2012discovering}, and controlling light pollution \cite{zhang2024causally}. Recent research also delves into the potential of urban imagery, employing remotely sensed images for poverty prediction, land cover classification \cite{jean2019tile2vec,hong2020graph,burke2021using}, and analyzing street view images to estimate pedestrian volume \cite{chen2020estimating}.

However, utilizing unimodal urban data often yields suboptimal results, prompting a growing inclination towards the integration of multi-modal data. For instance, the simultaneous utilization of streetscape and remote sensing imagery has proven effective in predicting socioeconomic indicators \cite{wang2018urban,li2022predicting}. The combination of urban imagery with POI data has demonstrated its utility in enhancing region representation \cite{wang2020urban2vec,huang2021m3g,liu2023knowledge}. Furthermore, researchers have ventured into the realm of multi-view graphs, leveraging data from diverse sources to comprehensively characterize regions \cite{qu2017attention,fu2019efficient}. This shift to multi-modal approaches holds great promise for advancing urban data analysis and interpretation, and helps to better achieve sustainable development goals. Recent efforts \cite{jean2019tile2vec,wu2022multi,liu2023knowledge} aim to derive latent embeddings for individual regions and employ them in conjunction with regional characteristics to predict a range of socioeconomic indicators, showcasing their notable versatility.

While prior studies have undertaken region profiling and socioeconomic prediction, several challenges persist. Among these, three primary ones emerge: (1) Rapid societal development has reshaped information exchange among regions, prompting a reassessment of the applicability of Tobler's First Law of Geography \cite{miller2004tobler}. Consequently, a more precise method is warranted to assess region similarity. (2) Urban representation predominantly focuses on geography and human activity, necessitating effective modal filtering to meet region representation demands amidst the abundance of urban data. (3) Achieving effective fusion of diverse modal data is crucial yet complex in developing the final region representation, necessitating the advancement of robust multi-modal fusion techniques.

To tackle these challenges, we present a Multi-Semantic Contrastive Learning (MuseCL) framework. The primary contributions of our work can be summarized as follows:

\begin{itemize}
    \item We pioneer the joint representation of regions using both street view and remote sensing imagery, concurrently integrating POI and mobility flow data to enrich the embedding with multi-dimensional semantic information.
    \item We enhance the spatial contrastive learning process by factoring in the similarity between regional POI and population mobility, resulting in more effective contrastive learning outcomes.
    \item We devise a cross-modal fusion model that aligns imagery with textual representation outputs, seamlessly integrating textual semantics into imagery representations.
    \item We validate the effectiveness of our framework through experiments on socioeconomic indicators in three major metropolises. The results demonstrate the superior performance of our model compared to various competitive state-of-the-art baselines across multiple downstream prediction tasks.
\end{itemize}

\begin{figure*}[t]
\centering
\includegraphics[width=0.99\textwidth]{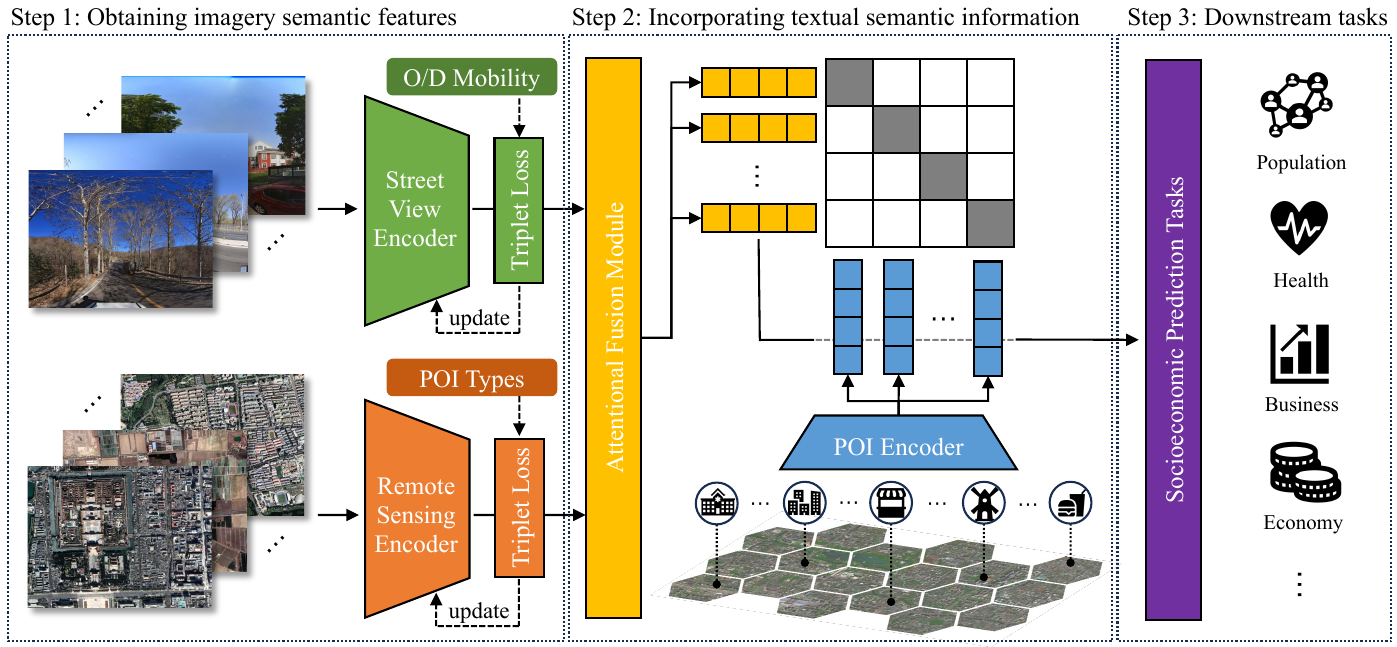}
\caption{The overall architecture of the proposed MuseCL.}
\label{framework}
\end{figure*}

\section{Related Work}
\paragraph{Urban Representation Learning.} With the increasing availability of urban data, representation learning in urban areas has witnessed significant growth in recent years. Numerous studies have capitalized on the proximity of similar regions in the embedding space to address various downstream tasks, such as crime prediction \cite{wang2017region,zhang2021multi}, land cover classification \cite{yao2018representing,luo2022urban}, and socioeconomic feature prediction \cite{wang2018urban,li2022predicting}, among others. In this context, various strategies have emerged for urban region representation. For instance, 
\citeauthor{feng2017poi2vec} \shortcite{feng2017poi2vec} proposed a latent representation model POI2Vec to jointly model the user preference and POI sequential transition influence for predicting potential visitors for a given POI. \citeauthor{wang2017region} \shortcite{wang2017region} introduced a method incorporating temporal dynamics and multi-hop transitions. \citeauthor{zhang2017regions} \shortcite{zhang2017regions} presented a novel cross-modal representation learning method, CrossMap, which uncovers urban dynamics with massive geo-tagged social media data. \citeauthor{yao2018representing} \shortcite{yao2018representing} proposed a framework to learn the vector representation of city zones by leveraging large-scale taxi trajectories. In a similar vein,  \citeauthor{fu2019efficient} \shortcite{fu2019efficient} explored multi-view spatial networks, considering geographical distance view and human mobility connectivity view for POIs within each region. Additionally, \citeauthor{wang2020urban2vec} \shortcite{wang2020urban2vec} devised a multi-modal and multi-stage framework integrating image and text data within the neighborhood.
These diverse approaches offer unique perspectives and valuable insights for further research in the urban representation learning field.

\paragraph{Socioeconomic Indicators Prediction.} Initially, researchers primarily employed supervised and unsupervised learning methods for predicting socioeconomic indicators. For instance, \citeauthor{chakraborty2016predicting} \shortcite{chakraborty2016predicting} proposed a generative model of real-world events to predict various socioeconomic indicators based on extracted events. \citeauthor{qu2017attention} \shortcite{qu2017attention} introduced a multi-view representation learning approach that fostered collaboration among different views to generate robust representations, 
subsequently used for socioeconomic indicator prediction. 
Similarly, \citeauthor{he2018perceiving} \shortcite{he2018perceiving} unveiled correlations between visual patterns in satellite images and commercial hotspots. In recent years, self-supervised learning methods, especially contrastive learning, have gained traction for socioeconomic indicator forecasting. Drawing inspiration from Tobler's First Law of Geography \cite{miller2004tobler}, \citeauthor{jean2019tile2vec} \shortcite{jean2019tile2vec} employed distance to establish neighborhood similarities in loss functions. Furthermore, \citeauthor{xi2022beyond} \shortcite{xi2022beyond} incorporated POI similarity into contrastive learning to overcome distance-based limitations.

\section{Preliminaries \& Problem Statement}
An urban area typically comprises multiple regions denoted as $\mathcal{R}=\{r_1, r_2, \cdots, r_N\}$. These regions exhibit unique geographic and demographic characteristics, often reflected through various data sources within them. In our study, we focus on analyzing specific attributes of regions $r_i \in \mathcal{R}$ ($i = 1, 2, \cdots, N$), investigating the following aspects:
\begin{itemize}
    \item \textbf{Remote Sensing Imagery $\mathcal{RV}_i$.} Remote sensing imagery captures ground surface details, effectively revealing building distribution and thus providing valuable support for region representation.
    \item \textbf{Street View Imagery $\mathcal{SV}_i=\{s_{i1}, s_{i2}, \cdots, s_{i |\mathcal{SV}_i|}\}$.} It offers valuable insights into the appearance of streets, buildings, and their immediate surroundings. A region often contains multiple street view images.
    \item \textbf{POI Data $\mathcal{T}_i=\{T_{i1}, T_{i2}, \cdots, T_{i |\mathcal{T}_i|}\}$.} We textualize each POI as a bag of words $\{t_1, t_2, \cdots, t_n\}$, where each word is obtained from the POI's categories, ratings, reviews, and other relevant information.
    \item \textbf{Population Mobility $\mathcal{M}_i=\{m^{in}_i, m^{out}_i\}$.} $m^{in}_i$ and $m^{out}_i$ refer to the number of people entering and exiting the region $r_i$ over a period of time, respectively. It can reflect the socio-demographic activity of a region.
\end{itemize}

Given a collection of urban remote sensing images $\mathcal{RV}$, street view images $\mathcal{SV}$, POI data $\mathcal{T}$, and population mobility data $\mathcal{M}$, our primary objective is to derive a low-dimensional representation $\epsilon_i \in \mathbb{R}^{d}$ for each region $r_i \in R (i = 1, 2, \cdots, N)$, where $d$ signifies the dimension of the representation vectors. By effectively encapsulating the diverse characteristics inherent in each region, our approach aims to generate compact yet informative representations, denoted as $\mathcal{E}=\{\epsilon_1, \epsilon_2, \cdots, \epsilon_N\}$, to enhance various downstream socioeconomic prediction tasks in urban settings.

\section{Methodology}
\subsection{Framework Overview}
Figure \ref{framework} illustrates our proposed framework for fine-grained urban region profiling to predict socioeconomic indicators. This multi-step contrastive learning model consists of three key components: extracting semantic features from the visual modality, incorporating textual semantic information, and performing downstream tasks.

To begin, we partition the visual semantic learning module into remote sensing imagery representations based on POI similarity and street view imagery representations based on population flow similarity. Contrastive learning sample pairs are curated to acquire imagery features with distinct focal points. Subsequently, we take into account the POI text information associated with each region and leverage a pre-trained encoder-based model to derive the text features for every region. Then, employing a feature-level attentive fusion module, we align the combined remote sensing and street view features with the text representation vectors of each region, thereby imbuing the fused features with both visual and textual semantic insights. Lastly, we evaluate the low-dimensional representations of each region across a range of downstream tasks critical for urban sustainable development.

\subsection{Visual Semantic Extraction}
Street view and remote sensing imagery often contain information with different emphases. For example, street view imagery can provide characteristics of the social environment and population activity, while remote sensing imagery is more oriented towards geographic attributes and surface features \cite{liu2023knowledge}. Therefore, we need to get the embedding of both separately and combine them effectively.

\subsubsection{Constructing Contrastive Samples}
\begin{figure}[t]
\centering
\includegraphics[width=1\columnwidth]{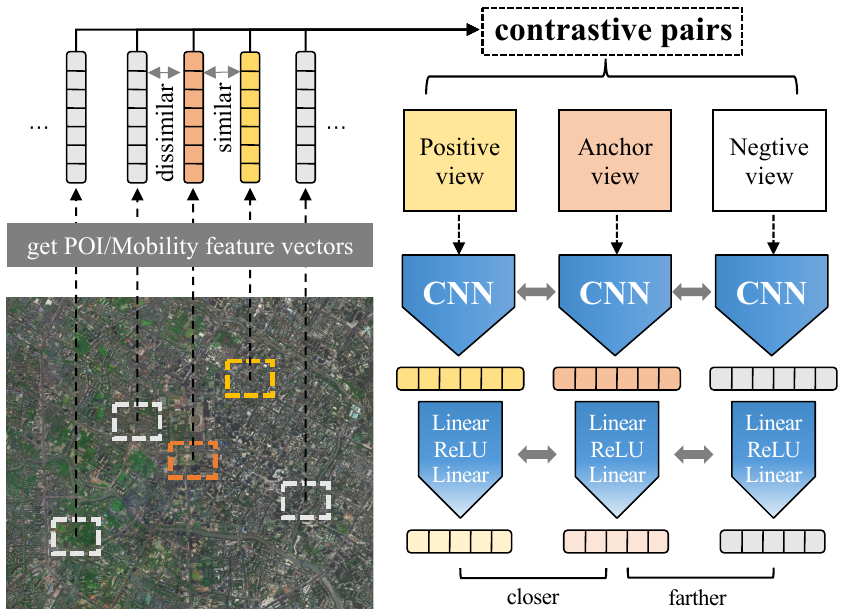}
\caption{Urban visual representation through contrastive learning based on POI and mobility similarities.}
\label{imagery_cl}
\end{figure}

Recently, \citeauthor{xi2022beyond} \shortcite{xi2022beyond} highlighted the limitations of Tobler’s First Law of Geography \cite{miller2004tobler}, noting that relying solely on spatial distance to measure regional similarity is flawed. To address this, we propose a refined approach by constructing contrastive learning pairs for street and remote sensing images based on population flow and POI similarity, respectively. Street images are paired with mobility data because they reflect human movement patterns, while remote sensing images are paired with POI data because they capture the built environment and land use.

Population flow within a region can be gauged by quantifying the influx and efflux of individuals or vehicles over a specified timeframe. If we conceptualize regions as nodes and the movement of individuals or vehicles between regions as edges, the mobility of each region can be captured by tallying the entries and exits at each node. Assuming that the inflow of population to region $r_i$ during a given period is $m_i^{in}$, and the outflow is $m_{i}^{out}$, the population mobility distance between regions $r_i$ and $r_j$ is computed by:

\begin{equation}
\label{dist_pm}
    dist_{i,j}^{\rm PM} = \sqrt{ \sum_{d \in \{in, out\}} \left( m^{d}_{i} - m^{d}_{j}\right)^2 }
\end{equation}
Then, the similarity of population mobility between the two regions can be quantified as:
\begin{equation}
    \lambda_{i,j}^{\rm PM} = \frac{1}{dist_{i,j}^{\rm PM}}
\end{equation}
We can construct positive samples characterized by higher similarity and negative samples characterized by lower similarity for each street view imagery, based on the parameter $\lambda_{i,j}^{\rm PM}$. As for remote sensing imagery, assuming that $K$ different POI types are considered, we employ the Euclidean distance to quantify the POI distance between region $r_i$ and $r_j$ as follows:
\begin{equation}
\label{dist_poi}
dist_{i,j}^{\rm POI} = \sqrt{\sum_{k=1}^{K}\left(POI_i^{k}-POI_j^{k} \right)^2}
\end{equation}
Therefore, the POI similarity between region $r_i$ and $r_j$ is:
\begin{equation}
    \lambda_{i,j}^{\rm POI} = \frac{1}{dist_{i,j}^{\rm POI}}
\end{equation}

\subsubsection{POI / Mobility Triplet Loss}
Using the acquired $\lambda_{i,j}^{\rm PM}$ and $\lambda_{i,j}^{\rm POI}$, we proceed to form separate pairs of contrastive learning samples for street view and remote sensing imagery. For instance, focusing on street view imagery, we establish each anchor image $Anc_i^{\rm SV}$ along with its corresponding positive sample $Pos_i^{\rm SV}$ and negative sample $Neg_i^{\rm SV}$ based on the population flow similarity $\lambda_{i,j}^{\rm PM}$. Subsequently, we train a convolutional neural network (CNN) denoted as $F_{\rm SV}$ to map the constructed contrastive learning samples $C^{\rm SV} = [Anc_i^{\rm SV}, Pos_i^{\rm SV}, Neg_i^{\rm SV}]$ into a low-dimensional vector space: $x_i^{\rm SV} = F_{\rm SV}(Anc_i^{\rm SV})$, $y_i^{\rm SV} = F_{\rm SV}(Pos_i^{\rm SV})$, and $z_i^{\rm SV} = F_{\rm SV}(Neg_i^{\rm SV})$. Similarly, we derive representation vectors for remote sensing imagery denoted as $x_i^{\rm RV}$, $y_i^{\rm RV}$, and $z_i^{\rm RV}$, corresponding to the contrastive learning samples $C^{\rm RV} = [Anc_i^{\rm RV}, Pos_i^{\rm RV}, Neg_i^{\rm RV}]$.

\subsubsection{Loss Optimization}
To ensure the minimization of the distance between the anchor image and the positive image, while maximizing the separation from the negative image in the representation space, we employ Triplet Loss \cite{schroff2015facenet} as the loss function. The primary objective of this loss function is to bring features with similar labels into close proximity within the representation space, while simultaneously pushing features with dissimilar labels apart. For each pair of samples, we anticipate the fulfillment of the following equations:
\begin{equation}
    {\rm sim}(x_i^m, y_i^m) + a \leq {\rm sim}(x_i^m, z_i^m), m \in \{{\rm SV},{\rm RV}\}
\end{equation}
\begin{equation}
    {\rm Loss}(C^{m}) = [a+{\rm sim}(x_i^m, y_i^m) - {\rm sim}(x_i^m, z_i^m)]_{+}
\end{equation}
where $[\cdot]_{+}$ is a rectifier function to keep the loss function value non-negative, and ${\rm sim}(\cdot)$ denotes the cosine similarity. The value $a$ is used to prevent the features of anchor samples $Anc_i^m$, positive samples $Pos_i^m$ and negative samples $Neg_i^m$ from aggregating into a small space. The whole training framework is shown in Figure \ref{imagery_cl}.


\subsection{Textual Semantic Incorporation}
POIs hold significance as data points denoting specific landmarks on a map, often signifying distinct geographic locations such as stores, restaurants, parks, and more in cities. The textual descriptions associated with POIs can effectively capture the geographic attributes of a region. For instance, a clustering of coffee shops within a region could indicate a vibrant locale appealing to young residents, whereas an abundance of parks and green spaces might suggest a neighborhood conducive to family-oriented living.

\subsubsection{POI Textual Semantic Extraction}
In addition to the imagery features, the textual data associated with POIs plays a crucial role in region profiling. To effectively harness the descriptive potential of POI text for region representation, we employ Gensim in conjunction with Skip-Gram and Huffman Softmax models \cite{mikolov2013efficient} for training. The Skip-Gram model, a neural network-based word vector approach, enables the learning of word vectors by predicting the contextual information of a word. Concurrently, the Huffman Softmax model, which leverages Huffman trees, enhances the neural network's output layer, refining the overall representation process.

Considering the complexity and ambiguity inherent in POI comments, we adopt a two-phase approach to extract the textual semantics. In the training phase, we utilize all POI comments and categories to train the model. However, as we transition to the representation phase, our focus narrows to utilizing solely the categorical information associated with each POI within the target regions.  Assuming that for region $r_i \in \mathcal{R}$, its POI data $\mathcal{T}_i=\{T_{i1}, T_{i2}, \cdots, T_{i |\mathcal{T}_i|}\}$ is the category of each POI in the region, and the final mapping of the trained model from words to vectors is $W$. Then the final POI embedding result for each region is:

\begin{equation}
    e_i^{\rm POI} = \frac{1}{|\mathcal{T}_i|} \sum_{j=1}^{|\mathcal{T}_i|} W(T_{ij}),T_{ij} \in \mathcal{T}_i
\end{equation}

\subsubsection{Attentive Fusion Module}
We proceed to integrate the street view features $e_i^{\rm SV}$, remote sensing features $e_i^{\rm RV}$, and POI features $e_i^{\rm POI}$, creating a comprehensive final representation tailored for utilization in various downstream tasks.

Firstly, with the inherent importance of both imagery features $e_i^{\rm SV}$ and $e_i^{\rm RV}$ unknown, we propose the incorporation of an attentive fusion module to derive weights for each of these representations. Considering street view features $e_i^{\rm SV}$ and remote sensing features $e_i^{\rm RV}$ from region $r_i$, we introduce learnable parameters $\mathbf{c}$, $\mathbf{M}$, and $\mathbf{b}$ to facilitate their fusion:
\begin{equation}
    \alpha_i^m = \mathbf{c}^T \cdot {\rm ReLU}(\mathbf{M} \cdot e_i^m + \mathbf{b}), m \in \{{\rm SV}, {\rm RV}\}
\end{equation}
\begin{equation}
    \beta_i^m = \frac{{\rm exp}(\alpha_i^m)}{\sum_{m \in \{{\rm SV}, {\rm RV}\}} {\rm exp}(\alpha_i^m)}
\end{equation}
\begin{equation}
    e_i^{\rm Image} = \sum_{m \in \{{\rm SV}, {\rm RV}\}} \beta_i^m \cdot e_i^m
\end{equation}
where $e_i^{\rm Image}$ is the final representation for region's imagery feature and $\beta_i^m$ ($m \in \{{\rm SV}, {\rm RV}\}$) are weight coefficients.

Next, in order to incorporate the textual semantic information of POIs, we refer to InfoNCE loss \cite{oord2018representation} to align the features of imagery $e_i^{\rm Image}$ and POI texts $e_i^{\rm POI}$:
\begin{equation}
    {\rm Loss}_i = - {\rm log} \frac{ {\rm exp}({\rm sim}(e_i^{\rm Image}, e_i^{\rm POI})) }{ \sum_{j=1}^{n} {\rm exp}({\rm sim}(e_i^{\rm Image}, e_j^{\rm POI})) }
\end{equation}
where $n$ denotes the mini-batch size. By optimizing the aforementioned loss function, we acquire the region imagery features $e_i^{\rm Image}$ and effectively integrate the semantic information of POIs. Subsequently, these obtained representations for each region can be harnessed to forecast various socioeconomic indicators.

\section{Experiments}
\subsection{Experimental Setups}
\subsubsection{Datasets}
We compile real-world datasets from three major cities: Beijing (BJ), Shanghai (SH), and New York (NY). The city regions are delineated by hexagonal divisions, with a radius of 1 km for Beijing and Shanghai, and 500 meters for New York (New York is much smaller than Beijing and Shanghai). It should be noted that our model is highly adaptable to various division shapes and scales, including road networks and Census Block Groups (CBGs).

For street view imagery, we employ the Baidu Maps API\footnote{https://lbsyun.baidu.com/index.php?title=viewstatic} for Beijing and Shanghai, and the Google Maps API\footnote{https://developers.google.com/maps} for New York. High-resolution (3.6-meter) remote sensing images are acquired through ArcGIS for all three cities. The POI data for Beijing and Shanghai originates from Baidu Maps, while New York's data is sourced from OpenStreetMap\footnote{https://www.openstreetmap.org/} (OSM). Socioeconomic indicators, including population density from WorldPop\footnote{https://hub.worldpop.org/}, housing data from Lianjia\footnote{https://m.lianjia.com/bj/ershoufang/index/}, and crime data from NYC Open Data\footnote{https://opendata.cityofnewyork.us/data/}, are also integrated.

\subsubsection{Baseline Models}
We compare our proposed model with various unimodal and multi-modal region representation algorithms, including:

\begin{itemize}
    \item \textbf{Inception v3} proposed in \cite{szegedy2016rethinking}. It can extract features using convolutional layers with different kernel sizes, max pooling and batch normalization.
    \item \textbf{Resnet-18} proposed in \cite{he2016deep}. It uses residual blocks to solve the degeneracy problem of deep networks. We use the Resnet-18 pre-trained in ImageNet.
    \item \textbf{Tile2vec} proposed in \cite{jean2019tile2vec}. It is an unsupervised learning method that uses geographic distance as a criterion for constructing contrastive samples.
    \item \textbf{Urban2vec} proposed in \cite{wang2020urban2vec}. It uses both street view images and POI data to characterize neighborhood features.
    \item \textbf{PG-SimCL} proposed in \cite{xi2022beyond}. It uses remote sensing imagery for region representation based on the similarity of geographic distances and POI distributions to perform prediction tasks on socioeconomic indicators.
    \item \textbf{Add-svrv} and \textbf{Fusion-svrv}. They represent the summation or attentional fusion of SV and RV embedding.
    \item \textbf{Concat}. We simply concatenate the SV, RV and POI representation results as a variant of our method.
\end{itemize}

\begin{table*}[t]
  \centering
  \resizebox{\textwidth}{!}{
    \begin{tabular}{c|cccccccccc}
    \toprule
    \textbf{City} & \multicolumn{10}{c}{\textbf{Beijing}} \\
    \midrule
    \multirow{2}[2]{*}{\textbf{Methods}} & \multicolumn{2}{c}{\textbf{PD}} & \multicolumn{2}{c}{\textbf{HD}} & \multicolumn{2}{c}{\textbf{MC}} & \multicolumn{2}{c}{\textbf{NP}} & \multicolumn{2}{c}{\textbf{NC}} \\
          & $R^2 \uparrow$    & $RMSE \downarrow$  & $R^2 \uparrow$    & $RMSE \downarrow$  & $R^2 \uparrow$    & $RMSE \downarrow$  & $R^2 \uparrow$    & $RMSE \downarrow$  & $R^2 \uparrow$    & $RMSE \downarrow$ \\
    \midrule
    Inception v3 & 0.0023  & 1.4765  & -0.0211  & 1.0325  & 0.0051  & 2.2586  & 0.0065  & 1.6991  & -0.0077  & 2.5665  \\
    Resnet-18 & 0.0685  & 1.4266  & -0.0235  & 1.0337  & 0.0344  & 2.2251  & 0.0356  & 1.6740  & 0.0263  & 2.5229  \\
    Tile2vec & 0.1102  & 1.3944  & -0.0071  & 1.0254  & 0.1104  & 2.1357  & 0.1137  & 1.6048  & 0.0951  & 2.4320  \\
    Urban2vec & 0.4982  & 1.0471  & 0.5327  & 0.6985  & \underline{0.5943}  & \underline{1.4422}  & \underline{0.7955}  & \underline{0.7708}  & 0.7251  & 1.3406  \\
    PG-SimCL & 0.1425  & 1.3688  & 0.1119  & 0.9630  & 0.1636  & 2.0708  & 0.1628  & 1.5597  & 0.1332  & 2.3804  \\
    Add-svrv & 0.0995 & 1.4027 & 0.0984 & 0.9702 & 0.0906 & 2.1593 & 0.1290 & 1.5909 & 0.1402 & 2.3707 \\
    Fusion-svrv & 0.1408 & 1.3701 & 0.1145 & 0.9615 & 0.1716 & 2.0609 & 0.1577 & 1.5645 & 0.1322 & 2.3817 \\
    Concat & \underline{0.4984}  & \underline{1.0469}  & \underline{0.5399}  & \underline{0.6931}  & 0.5738  & 1.4783  & 0.7832  & 0.7937  & \underline{0.7276}  & \underline{1.3345}  \\
    \midrule
    \textbf{Ours} & \textbf{0.5310} & \textbf{1.0123} & \textbf{0.5708} & \textbf{0.6694} & \textbf{0.6229} & \textbf{1.3906} & \textbf{0.9471} & \textbf{0.3921} & \textbf{0.8782} & \textbf{0.8921} \\
    Impr. & 6.54\% & 3.30\% & 5.72\% & 3.42\% & 4.81\% & 3.58\% & 19.06\% & 49.13\% & 20.70\% & 33.15\% \\
    \bottomrule
    \end{tabular}}
    \caption{Prediction results of different socioeconomic indicators for Beijing: Population Density (\textbf{PD}), Housing Density (\textbf{HD}), Mobility Count (\textbf{MC}), Number of POIs (\textbf{NP}) and Number of Comments (\textbf{NC}). The best results are \textbf{in bold} and the second best results are \underline{underlined}.}
  \label{beijing}
\end{table*}

\begin{figure*}[t]
\vspace{-0.8em}
\centering
\subfloat[Population Density]{
    \includegraphics[width=3.4cm]{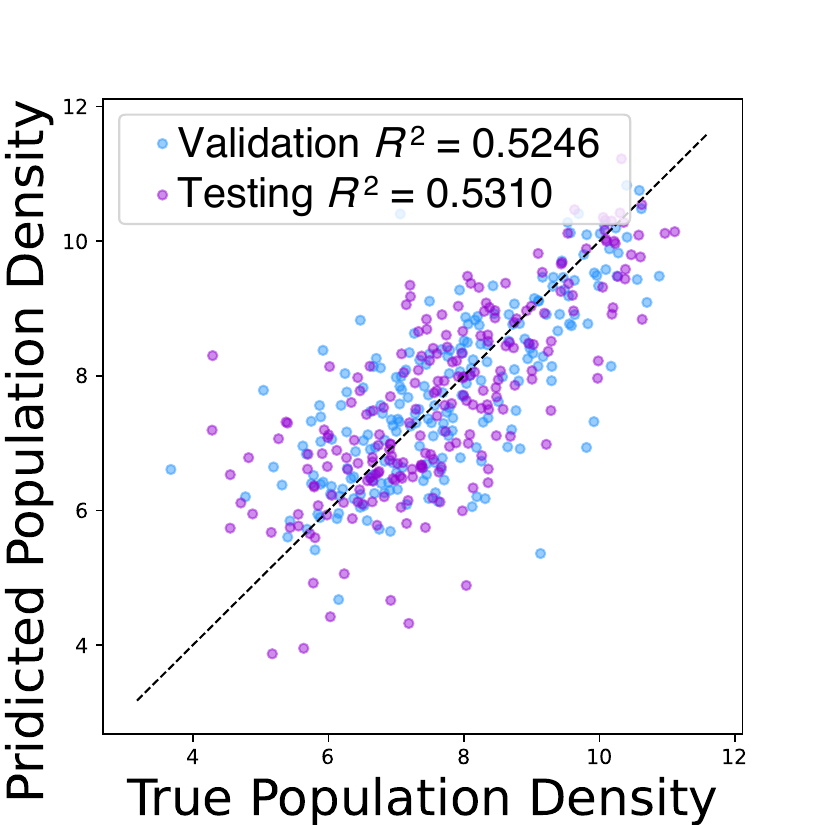}
}
\subfloat[Housing Density]{
    \includegraphics[width=3.4cm]{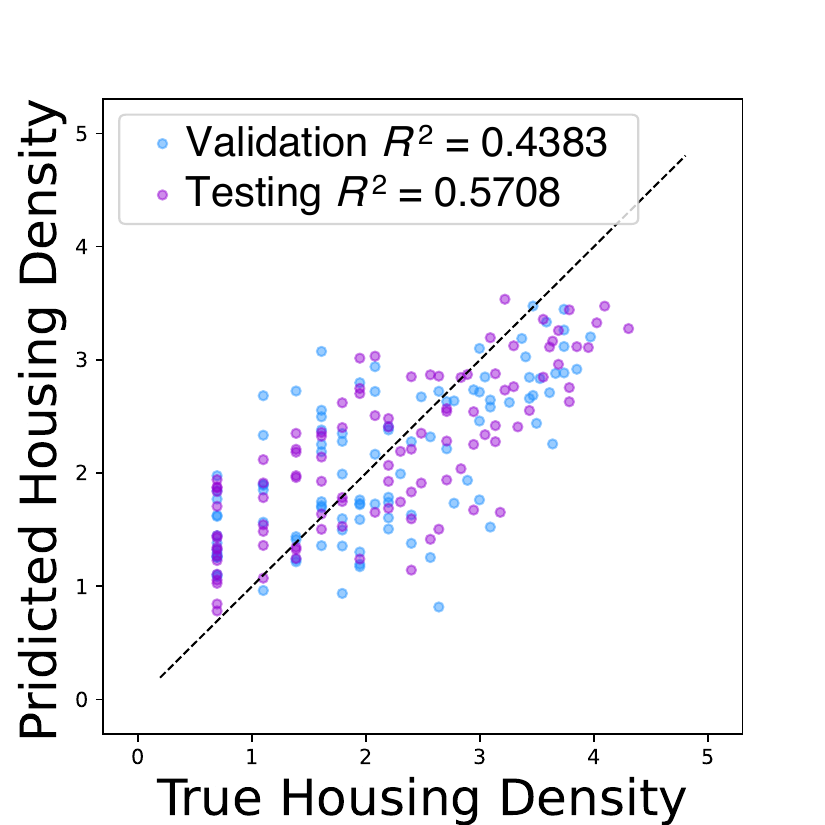}
}
\subfloat[Mobility Count]{
    \includegraphics[width=3.4cm]{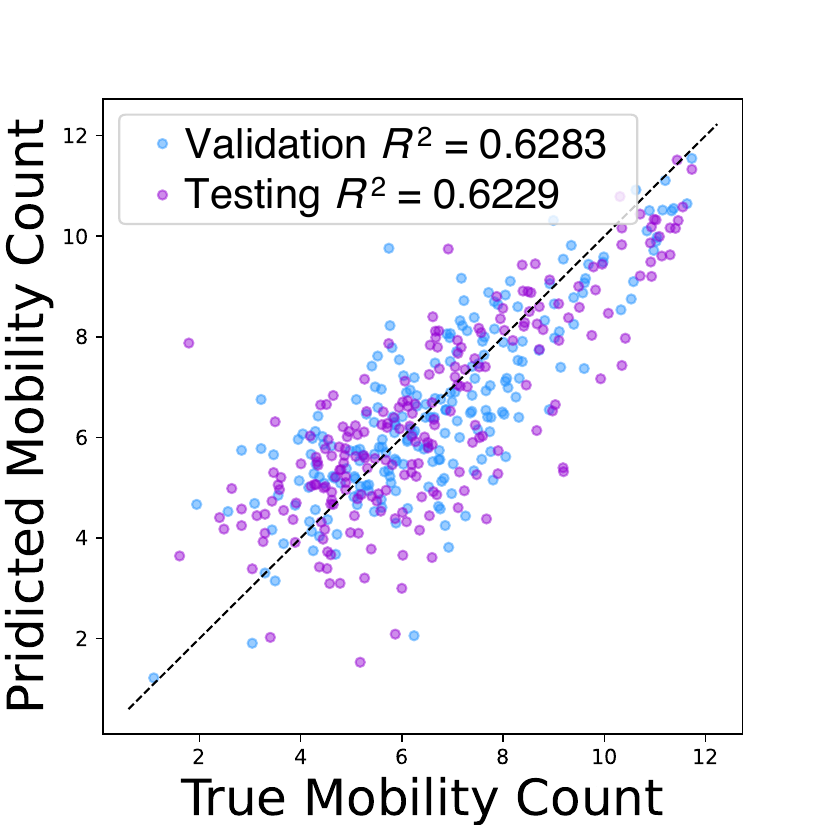}
}
\subfloat[Number of POIs]{
    \includegraphics[width=3.4cm]{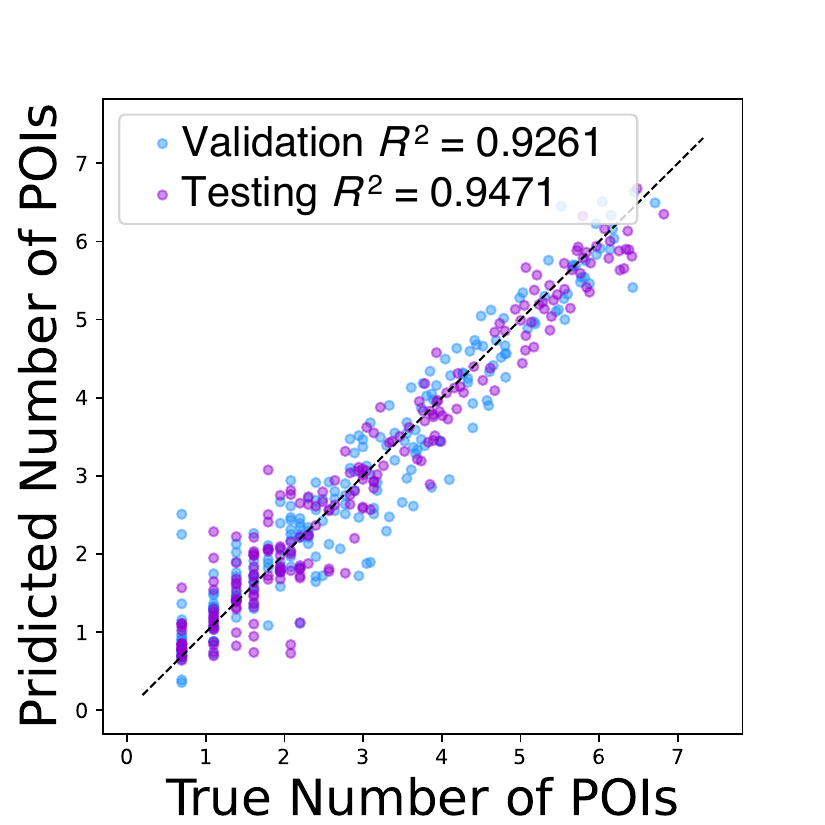}
}
\subfloat[Number of Comments]{
    \includegraphics[width=3.4cm]{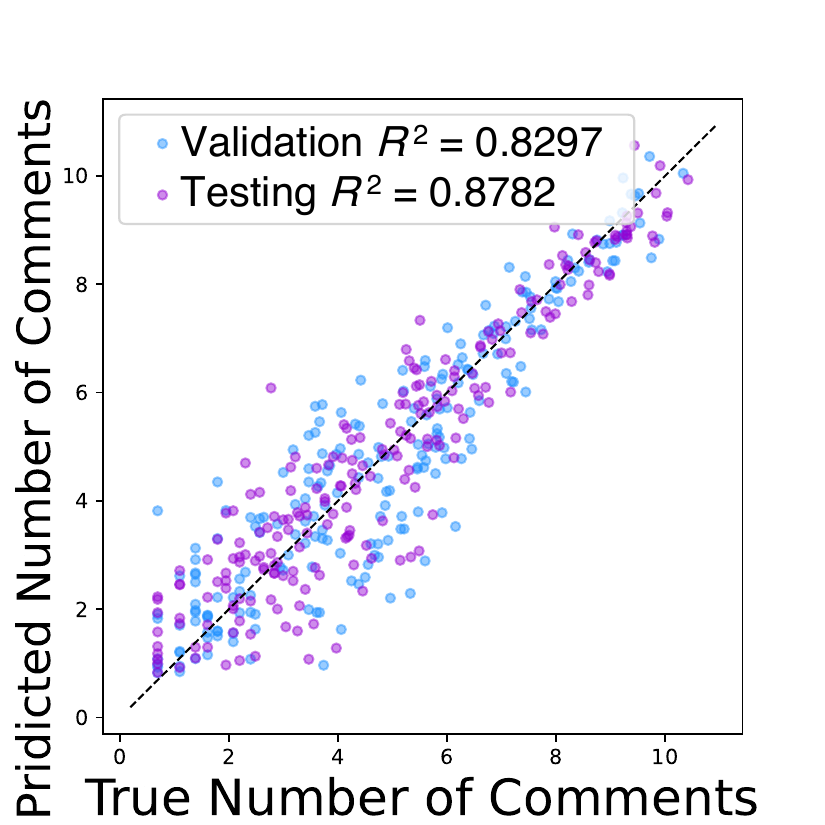}
}
 \caption{Predicted values versus true values on five socioeconomic datasets in Beijing. The dotted line is 45 degrees. The blue dots represent the regions of the validation set and the purple dots represent the regions of the test set with the respective $R^2$ illustrated.}
\label{prid_vs_true1}
\end{figure*}

\subsubsection{Metrics and Implementation}
We adopt rooted mean squared error ($RMSE$) and coefficient of determination ($R^2$) for evaluation. In our experiments, we use Inception v3 as SV encoder's backbone and Resnet-18 as RV encoder's backbone with a final linear layer projecting features into the 128-dimensional embedding space. We set the batch size to 32, the learning rate to 5$e^{-4}$, and use Adam optimizer. Datasets are split into 60\% training, 20\% validation, and 20\% test sets. All socioeconomic indicators are converted into logarithmic scale.

\subsection{Experimental Results}
\subsubsection{Socioeconomic Indicators Prediction}
We utilize the embedding of each region as input and employ a multi-layer perceptron (MLP) to predict the socioeconomic indicators. Table \ref{beijing} shows the prediction outcomes for Beijing, demonstrating a significant $R^2$ improvement of 4.81\% to 20.70\% over the best baselines across the five datasets.

Specifically, the \textbf{Inception v3} and \textbf{Resnet-18} networks pre-trained on ImageNet achieve average $R^2$ scores of only -0.0030 and 0.0283 across datasets, failing to adequately capture the relationships between regions. While \textbf{Tile2vec} and \textbf{PG-SimCL} show some improvements over pre-trained models with an average $R^2$ of 0.0845 and 0.1428, they both fall short in providing a comprehensive representation solely through remote sensing imagery. Then, \textbf{Urban2vec}, a multi-modal approach, outperforms other unimodal models with an average $R^2$ of 0.6292 exhibiting enhanced prediction results by incorporating street view images and POI data. A comparison of \textbf{Add-svrv} and \textbf{Fusion-svrv} reveals that the use of attentional fusion module is more effective than simple summation when visual semantic is utilized. Our \textbf{MuseCL} notably excels in all Beijing datasets with an average $R^2$ of 0.7100, which is 13.67\% higher than \textbf{Concat}, highlighting successful multi-semantic integration across visual and textual modalities. This also demonstrates the effectiveness of MuseCL in representing regional attributes, leading to more precise predictions of socioeconomic indicators. Furthermore,  Figure \ref{prid_vs_true1} shows the predicted value v.s. true value on socioeconomic indicators for Beijing, indicating that our MuseCL framework shows superior prediction effect on different indicators.

\begin{table*}[t]
  \centering
  \resizebox{\textwidth}{!}{
    \begin{tabular}{ccc|cccccc|cc}
    \toprule
    \textbf{City} & \textbf{Dataset} & Metrics & Inception v3 & Resnet-18 & Tile2vec & Urban2vec & PG-SimCL &Concat & \textbf{Ours} & Impr. \\
    \midrule
    \multirow{6}[6]{*}{\textbf{Shanghai}} & \multirow{2}[2]{*}{\textbf{PD}} & $R^2 \uparrow$    & -0.0384  & -0.2813  & 0.0016  & 0.3401 & 0.0261 & \underline{0.3596}  & \textbf{0.4301} & 19.61\% \\
          &       & $RMSE \downarrow$  & 1.2269  & 1.3629 & 1.2031  & 0.9780 & 1.1882 & \underline{0.9635}  & \textbf{0.9089} & 5.67\% \\
\cmidrule{2-11}          & \multirow{2}[2]{*}{\textbf{HD}} & $R^2 \uparrow$    & -0.0962  & -0.0889  & 0.0424  & \underline{0.4061} & -0.0245 & 0.3763  & \textbf{0.4330} & 6.62\% \\
          &       & $RMSE \downarrow$  & 1.0586  & 1.0551  & 0.9895  & \underline{0.7792} & 1.0234 & 0.7985  & \textbf{0.7614} & 2.28\% \\          
\cmidrule{2-11}          & \multirow{2}[2]{*}{\textbf{NP}} & $R^2 \uparrow$    & -0.0069  & -0.0540  & 0.0677  & \underline{0.8726} & 0.0902 & 0.8191  & \textbf{0.9283} & 6.38\% \\
          &       & $RMSE \downarrow$  & 1.5706  & 1.6069  & 1.5113  & \underline{0.5586} & 1.4930 & 0.6657  & \textbf{0.4191} & 24.97\% \\          
    \midrule
    \multirow{6}[6]{*}{\textbf{New York}} & \multirow{2}[2]{*}{\textbf{PD}} & $R^2 \uparrow$    & -0.0063  & -0.0042  & 0.0052  & \underline{0.3551} & 0.2735 & 0.3436  & \textbf{0.4165} & 17.29\% \\    
          &       & $RMSE \downarrow$  & 2.1691  & 2.1669  & 2.1568  & \underline{1.7365} & 1.8431 & 1.7519  & \textbf{1.6517} & 4.88\% \\          
\cmidrule{2-11}          & \multirow{2}[2]{*}{\textbf{CR}} & $R^2 \uparrow$    & -0.0001  & -0.0146  & -0.0442  & \underline{0.3183} & 0.1634 & 0.2979  & \textbf{0.3303} & 3.77\% \\
          &       & $RMSE \downarrow$  & 1.5713  & 1.5827  & 1.6056  & \underline{1.2973} & 1.4371 & 1.3166  & \textbf{1.2858} & 0.89\% \\          
\cmidrule{2-11}          & \multirow{2}[2]{*}{\textbf{NP}} & $R^2 \uparrow$    & -0.0416  & -0.0531  & -0.0117  & \underline{0.2397} & 0.2102 & 0.2395  & \textbf{0.2594} & 8.22\% \\
          &       & $RMSE \downarrow$  & 1.3963  & 1.4040  & 1.3761  & \underline{1.1929} & 1.2159 & 1.1931  & \textbf{1.1774} & 1.30\% \\         
    \bottomrule
    \end{tabular}}
    \caption{Prediction results of different socioeconomic indicators for Shanghai and New York: Population Density (\textbf{PD}), Housing Density (\textbf{HD}), Number of POIs (\textbf{NP}) and Crime (\textbf{CR}). The best results are \textbf{in bold} and the second best results are \underline{underlined}.}
  \label{SH/NY}
\end{table*}

\subsubsection{Model Adaptability to Other Cities}
We expand our experimental scope to include other well-developed cities, thereby testing the adaptability of our model. Shanghai has 746 valid representation regions, while New York has 517. In Shanghai, our experiments cover Population Density (\textbf{PD}), Housing Density (\textbf{HD}), and Number of POIs (\textbf{NP}) indicators. For New York, we incorporate the widely recognized Crime (\textbf{CR}) dataset. Consistency in comparisons is maintained by employing the same baseline models as before. The predictive outcomes are presented in Table \ref{SH/NY}. Notably, our model consistently outperforms across cities with varying sizes and geographical characteristics. Compared to the next-best baselines, we achieve an improvement in $R^2$ ranging from 3.77\% to 19.61\%. This robust performance reaffirms the adaptability of our model in addressing the demands of different city types and diverse datasets.

\subsubsection{Visualization of Region Representations}

\begin{figure}[t]
\centering
\includegraphics[width=1\columnwidth]{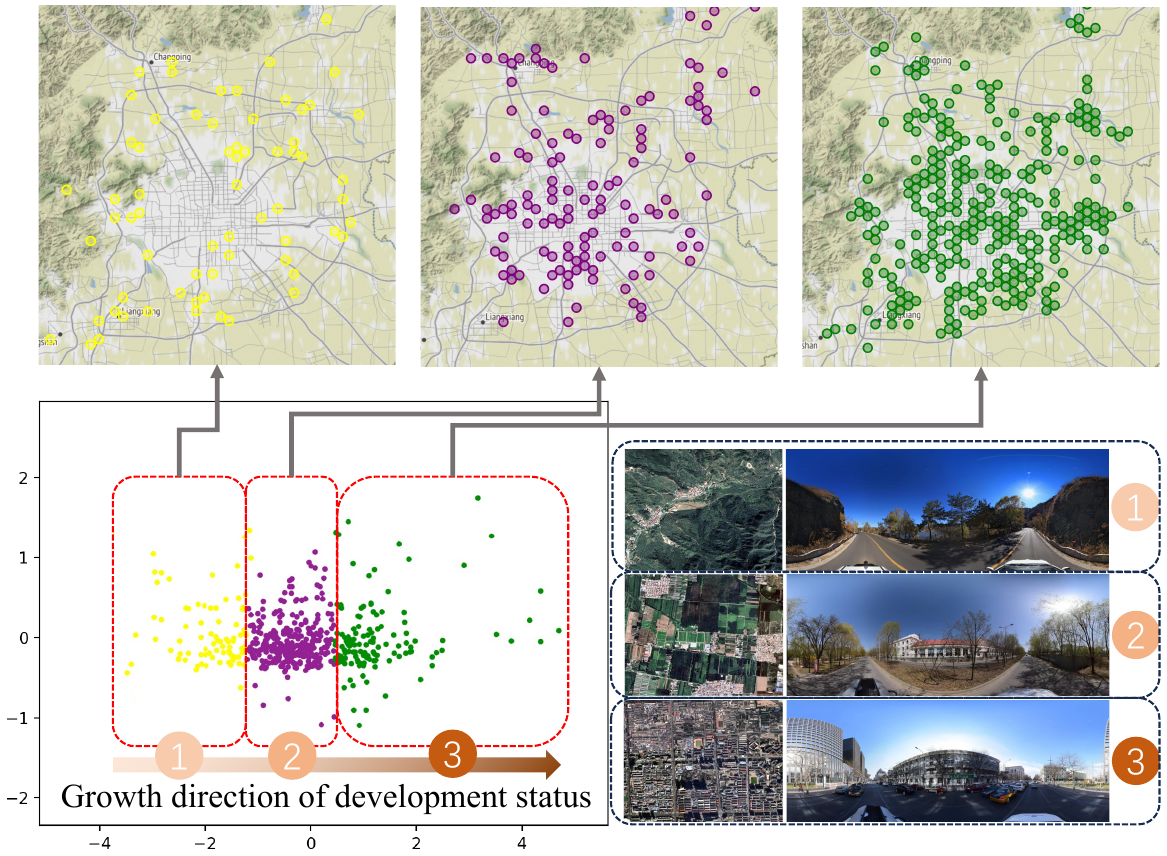}
\caption{Visualization of the final representation space.}
\label{visual}
\end{figure}

To gain deeper insights into the embedding space, we apply principal components analysis (PCA) \cite{shlens2014tutorial} to downscale the representation vectors. We then use the K-means algorithm to classify the regions into three clusters and depict their spatial distribution in Figure \ref{visual}. This visualization reveals that regions with different levels of development occupy distinct locations in the embedded space.

Specifically, the yellow regions are situated on the outskirts of Beijing, indicating underdevelopment and low socioeconomic attributes. Their remote sensing and street view images depict agricultural landscapes with sparse populations and limited POIs. In addition, purple regions, which are moderately developed, extend across urban and suburban areas, revealing emerging villages in their imagery. These areas maintain non-built spaces but exhibit higher population densities and POI counts compared to the yellow ones. Meanwhile, green regions, in central urban zones, include Beijing's commercial hubs, showing a highly urbanized environment with strong socioeconomic indicators.

\subsubsection{Ablation Study}
\begin{figure}[t]
\vspace{-0.9em}
\centering
\subfloat[Beijing]{
    \includegraphics[width=4cm]{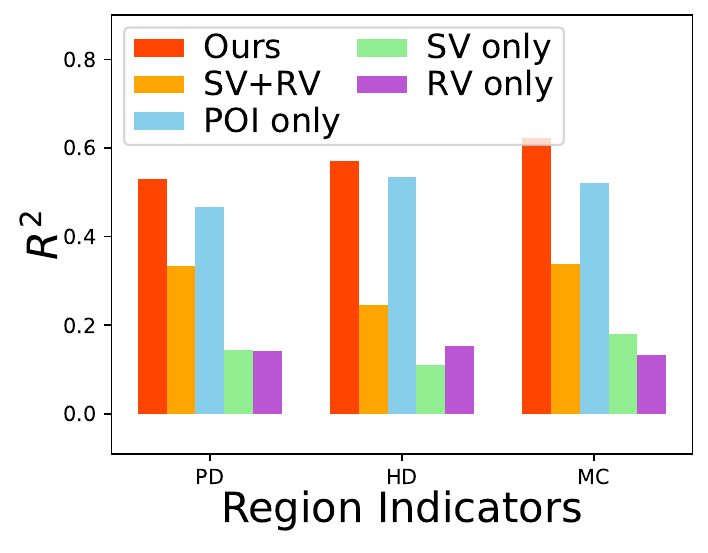}
}
\subfloat[New York]{
    \includegraphics[width=4cm]{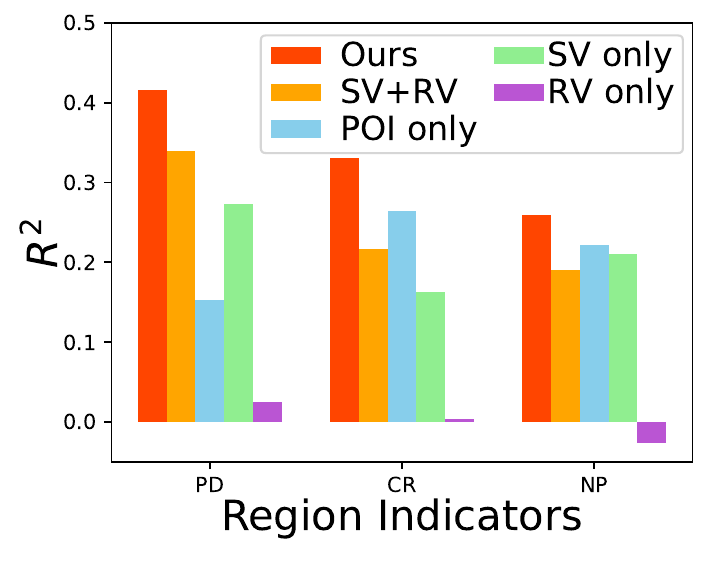}
}
\caption{Results of ablation study in Beijing and New York.}
\label{ablation}
\end{figure}

We conduct ablation experiments using three datasets each from Beijing and New York City. The results, as depicted in Figure \ref{ablation}, indicate that the absence of certain modalities leads to a reduction in the final prediction $R^2$ value. Notably, relying solely on POI, street view, or remote sensing images yields suboptimal outcomes. When combining street view and remote sensing images without POI information, the performance still falls short of our model's performance, although it fares better than utilizing street view or remote sensing images individually. This reinforces the notion that various modalities contribute distinct insights for predicting downstream tasks and urban region profiling.

\section{Conclusion}
This paper presents a novel Multi-Semantic Contrastive Learning (MuseCL) framework that skillfully amalgamates semantic insights from visual and textual information to generate embeddings for urban regions. We showcase our model's superiority in socioeconomic indicators prediction across diverse cities and through extended experiments. While our focus is on statically depicting urban regions, it is important to acknowledge their rapid evolution due to development. Therefore, incorporating time into region representation presents an interesting path for future research.

\section*{Acknowledgments}
This work was supported by Natural Science Foundation of China (NSFC Grant No.~62106274) and the Fundamental Research Funds for the Central Universities, Renmin University of China (Grant No.~22XNKJ24). We also wish to acknowledge the support provided by Engineering Research Center of Next-Generation Intelligent Search and Recommendation, Ministry of Education and the Intelligent Social Governance Platform, Major Innovation \& Planning Interdisciplinary Platform for the ``Double-First Class'' Initiative, Renmin University of China.

\bibliographystyle{named}
\bibliography{ijcai24}

\begin{thebibliography}{}

\bibitem[\protect\citeauthoryear{Aghababaei and Makrehchi}{2016}]{aghababaei2016mining}
Somayyeh Aghababaei and Masoud Makrehchi.
\newblock Mining social media content for crime prediction.
\newblock In {\em 2016 IEEE/WIC/ACM International Conference on Web Intelligence (WI)}, pages 526--531. IEEE, 2016.

\bibitem[\protect\citeauthoryear{Antenucci \bgroup \em et al.\egroup }{2014}]{antenucci2014using}
Dolan Antenucci, Michael Cafarella, Margaret Levenstein, Christopher R{\'e}, and Matthew~D Shapiro.
\newblock Using social media to measure labor market flows.
\newblock Technical report, National Bureau of Economic Research, 2014.

\bibitem[\protect\citeauthoryear{Burke \bgroup \em et al.\egroup }{2021}]{burke2021using}
Marshall Burke, Anne Driscoll, David~B Lobell, and Stefano Ermon.
\newblock Using satellite imagery to understand and promote sustainable development.
\newblock {\em Science}, 371(6535):eabe8628, 2021.

\bibitem[\protect\citeauthoryear{Chakraborty \bgroup \em et al.\egroup }{2016}]{chakraborty2016predicting}
Sunandan Chakraborty, Ashwin Venkataraman, Srikanth Jagabathula, and Lakshminarayanan Subramanian.
\newblock Predicting socio-economic indicators using news events.
\newblock In {\em Proceedings of the 22nd ACM SIGKDD international conference on knowledge discovery and data mining}, pages 1455--1464, 2016.

\bibitem[\protect\citeauthoryear{Chen \bgroup \em et al.\egroup }{2020}]{chen2020estimating}
Long Chen, Yi~Lu, Qiang Sheng, Yu~Ye, Ruoyu Wang, and Ye~Liu.
\newblock Estimating pedestrian volume using street view images: A large-scale validation test.
\newblock {\em Computers, Environment and Urban Systems}, 81:101481, 2020.

\bibitem[\protect\citeauthoryear{Cohen \bgroup \em et al.\egroup }{2016}]{cohen2016using}
Peter Cohen, Robert Hahn, Jonathan Hall, Steven Levitt, and Robert Metcalfe.
\newblock Using big data to estimate consumer surplus: The case of uber.
\newblock Technical report, National Bureau of Economic Research, 2016.

\bibitem[\protect\citeauthoryear{Custodio \bgroup \em et al.\egroup }{2023}]{custodio2023review}
Henry~M Custodio, Michalis Hadjikakou, and Brett~A Bryan.
\newblock A review of socioeconomic indicators of sustainability and wellbeing building on the social foundations framework.
\newblock {\em Ecological Economics}, 203:107608, 2023.

\bibitem[\protect\citeauthoryear{Feng \bgroup \em et al.\egroup }{2017}]{feng2017poi2vec}
Shanshan Feng, Gao Cong, Bo~An, and Yeow~Meng Chee.
\newblock Poi2vec: Geographical latent representation for predicting future visitors.
\newblock In {\em Proceedings of the AAAI Conference on Artificial Intelligence}, volume~31, 2017.

\bibitem[\protect\citeauthoryear{Fu \bgroup \em et al.\egroup }{2019}]{fu2019efficient}
Yanjie Fu, Pengyang Wang, Jiadi Du, Le~Wu, and Xiaolin Li.
\newblock Efficient region embedding with multi-view spatial networks: A perspective of locality-constrained spatial autocorrelations.
\newblock In {\em Proceedings of the AAAI Conference on Artificial Intelligence}, volume~33, pages 906--913, 2019.

\bibitem[\protect\citeauthoryear{Gao \bgroup \em et al.\egroup }{2017}]{gao2017identifying}
Qiang Gao, Fan Zhou, Kunpeng Zhang, Goce Trajcevski, Xucheng Luo, and Fengli Zhang.
\newblock Identifying human mobility via trajectory embeddings.
\newblock In {\em IJCAI}, volume~17, pages 1689--1695, 2017.

\bibitem[\protect\citeauthoryear{Habitat}{2022}]{habitat2022world}
UN~Habitat.
\newblock World cities report 2022: Envisaging the future of cities.
\newblock {\em United Nations Human Settlements Programme: Nairobi, Kenya}, pages 41--44, 2022.

\bibitem[\protect\citeauthoryear{He \bgroup \em et al.\egroup }{2016}]{he2016deep}
Kaiming He, Xiangyu Zhang, Shaoqing Ren, and Jian Sun.
\newblock Deep residual learning for image recognition.
\newblock In {\em Proceedings of the IEEE conference on computer vision and pattern recognition}, pages 770--778, 2016.

\bibitem[\protect\citeauthoryear{He \bgroup \em et al.\egroup }{2018}]{he2018perceiving}
Zhiyuan He, Su~Yang, Weishan Zhang, and Jiulong Zhang.
\newblock Perceiving commerial activeness over satellite images.
\newblock In {\em Companion Proceedings of the The Web Conference 2018}, pages 387--394, 2018.

\bibitem[\protect\citeauthoryear{Hong \bgroup \em et al.\egroup }{2020}]{hong2020graph}
Danfeng Hong, Lianru Gao, Jing Yao, Bing Zhang, Antonio Plaza, and Jocelyn Chanussot.
\newblock Graph convolutional networks for hyperspectral image classification.
\newblock {\em IEEE Transactions on Geoscience and Remote Sensing}, 59(7):5966--5978, 2020.

\bibitem[\protect\citeauthoryear{Huang \bgroup \em et al.\egroup }{2021}]{huang2021m3g}
Tianyuan Huang, Zhecheng Wang, Hao Sheng, Andrew~Y Ng, and Ram Rajagopal.
\newblock M3g: Learning urban neighborhood representation from multi-modal multi-graph.
\newblock In {\em Proceedings of the DeepSpatial 2021: 2nd ACM KDD Workshop on Deep Learning for Spatio-Temporal Data, Applications and Systems}, 2021.

\bibitem[\protect\citeauthoryear{Ilieva and McPhearson}{2018}]{ilieva2018social}
Rositsa~T Ilieva and Timon McPhearson.
\newblock Social-media data for urban sustainability.
\newblock {\em Nature Sustainability}, 1(10):553--565, 2018.

\bibitem[\protect\citeauthoryear{Jean \bgroup \em et al.\egroup }{2019}]{jean2019tile2vec}
Neal Jean, Sherrie Wang, Anshul Samar, George Azzari, David Lobell, and Stefano Ermon.
\newblock Tile2vec: Unsupervised representation learning for spatially distributed data.
\newblock In {\em Proceedings of the AAAI Conference on Artificial Intelligence}, volume~33, pages 3967--3974, 2019.

\bibitem[\protect\citeauthoryear{Li \bgroup \em et al.\egroup }{2022}]{li2022predicting}
Tong Li, Shiduo Xin, Yanxin Xi, Sasu Tarkoma, Pan Hui, and Yong Li.
\newblock Predicting multi-level socioeconomic indicators from structural urban imagery.
\newblock In {\em Proceedings of the 31st ACM International Conference on Information \& Knowledge Management}, pages 3282--3291, 2022.

\bibitem[\protect\citeauthoryear{Liu \bgroup \em et al.\egroup }{2023}]{liu2023knowledge}
Yu~Liu, Xin Zhang, Jingtao Ding, Yanxin Xi, and Yong Li.
\newblock Knowledge-infused contrastive learning for urban imagery-based socioeconomic prediction.
\newblock In {\em Proceedings of the ACM Web Conference 2023}, pages 4150--4160, 2023.

\bibitem[\protect\citeauthoryear{Luo \bgroup \em et al.\egroup }{2022}]{luo2022urban}
Yan Luo, Fu-lai Chung, and Kai Chen.
\newblock Urban region profiling via multi-graph representation learning.
\newblock In {\em Proceedings of the 31st ACM International Conference on Information \& Knowledge Management}, pages 4294--4298, 2022.

\bibitem[\protect\citeauthoryear{Mikolov \bgroup \em et al.\egroup }{2013}]{mikolov2013efficient}
Tomas Mikolov, Kai Chen, Greg Corrado, and Jeffrey Dean.
\newblock Efficient estimation of word representations in vector space.
\newblock {\em arXiv preprint arXiv:1301.3781}, 2013.

\bibitem[\protect\citeauthoryear{Miller}{2004}]{miller2004tobler}
Harvey~J Miller.
\newblock Tobler's first law and spatial analysis.
\newblock {\em Annals of the association of American geographers}, 94(2):284--289, 2004.

\bibitem[\protect\citeauthoryear{Oord \bgroup \em et al.\egroup }{2018}]{oord2018representation}
Aaron van~den Oord, Yazhe Li, and Oriol Vinyals.
\newblock Representation learning with contrastive predictive coding.
\newblock {\em arXiv preprint arXiv:1807.03748}, 2018.

\bibitem[\protect\citeauthoryear{Qu \bgroup \em et al.\egroup }{2017}]{qu2017attention}
Meng Qu, Jian Tang, Jingbo Shang, Xiang Ren, Ming Zhang, and Jiawei Han.
\newblock An attention-based collaboration framework for multi-view network representation learning.
\newblock In {\em Proceedings of the 2017 ACM on Conference on Information and Knowledge Management}, pages 1767--1776, 2017.

\bibitem[\protect\citeauthoryear{Sachs \bgroup \em et al.\egroup }{2022}]{sachs2022sustainable}
Jeffrey~D Sachs, Christian Kroll, Guillame Lafortune, Grayson Fuller, and Finn Woelm.
\newblock {\em Sustainable development report 2022}.
\newblock Cambridge University Press, 2022.

\bibitem[\protect\citeauthoryear{Schroff \bgroup \em et al.\egroup }{2015}]{schroff2015facenet}
Florian Schroff, Dmitry Kalenichenko, and James Philbin.
\newblock Facenet: A unified embedding for face recognition and clustering.
\newblock In {\em Proceedings of the IEEE conference on computer vision and pattern recognition}, pages 815--823, 2015.

\bibitem[\protect\citeauthoryear{Shlens}{2014}]{shlens2014tutorial}
Jonathon Shlens.
\newblock A tutorial on principal component analysis.
\newblock {\em arXiv preprint arXiv:1404.1100}, 2014.

\bibitem[\protect\citeauthoryear{Szegedy \bgroup \em et al.\egroup }{2016}]{szegedy2016rethinking}
Christian Szegedy, Vincent Vanhoucke, Sergey Ioffe, Jon Shlens, and Zbigniew Wojna.
\newblock Rethinking the inception architecture for computer vision.
\newblock In {\em Proceedings of the IEEE conference on computer vision and pattern recognition}, pages 2818--2826, 2016.

\bibitem[\protect\citeauthoryear{Wang and Li}{2017}]{wang2017region}
Hongjian Wang and Zhenhui Li.
\newblock Region representation learning via mobility flow.
\newblock In {\em Proceedings of the 2017 ACM on Conference on Information and Knowledge Management}, pages 237--246, 2017.

\bibitem[\protect\citeauthoryear{Wang \bgroup \em et al.\egroup }{2018a}]{wang2018urban}
Wenshan Wang, Su~Yang, Zhiyuan He, Minjie Wang, Jiulong Zhang, and Weishan Zhang.
\newblock Urban perception of commercial activeness from satellite images and streetscapes.
\newblock In {\em Companion Proceedings of the The Web Conference 2018}, pages 647--654, 2018.

\bibitem[\protect\citeauthoryear{Wang \bgroup \em et al.\egroup }{2018b}]{wang2018predicting}
Yingzi Wang, Xiao Zhou, Cecilia Mascolo, Anastasios Noulas, Xing Xie, and Qi~Liu.
\newblock Predicting the spatio-temporal evolution of chronic diseases in population with human mobility data.
\newblock In {\em IJCAI}, 2018.

\bibitem[\protect\citeauthoryear{Wang \bgroup \em et al.\egroup }{2020}]{wang2020urban2vec}
Zhecheng Wang, Haoyuan Li, and Ram Rajagopal.
\newblock Urban2vec: Incorporating street view imagery and pois for multi-modal urban neighborhood embedding.
\newblock In {\em Proceedings of the AAAI Conference on Artificial Intelligence}, volume~34, pages 1013--1020, 2020.

\bibitem[\protect\citeauthoryear{Wu \bgroup \em et al.\egroup }{2022}]{wu2022multi}
Shangbin Wu, Xu~Yan, Xiaoliang Fan, Shirui Pan, Shichao Zhu, Chuanpan Zheng, Ming Cheng, and Cheng Wang.
\newblock Multi-graph fusion networks for urban region embedding.
\newblock {\em arXiv preprint arXiv:2201.09760}, 2022.

\bibitem[\protect\citeauthoryear{Xi \bgroup \em et al.\egroup }{2022}]{xi2022beyond}
Yanxin Xi, Tong Li, Huandong Wang, Yong Li, Sasu Tarkoma, and Pan Hui.
\newblock Beyond the first law of geography: Learning representations of satellite imagery by leveraging point-of-interests.
\newblock In {\em Proceedings of the ACM Web Conference 2022}, pages 3308--3316, 2022.

\bibitem[\protect\citeauthoryear{Yao \bgroup \em et al.\egroup }{2018}]{yao2018representing}
Zijun Yao, Yanjie Fu, Bin Liu, Wangsu Hu, and Hui Xiong.
\newblock Representing urban functions through zone embedding with human mobility patterns.
\newblock In {\em Proceedings of the Twenty-Seventh International Joint Conference on Artificial Intelligence (IJCAI-18)}, 2018.

\bibitem[\protect\citeauthoryear{Yuan \bgroup \em et al.\egroup }{2012}]{yuan2012discovering}
Jing Yuan, Yu~Zheng, and Xing Xie.
\newblock Discovering regions of different functions in a city using human mobility and pois.
\newblock In {\em Proceedings of the 18th ACM SIGKDD international conference on Knowledge discovery and data mining}, pages 186--194, 2012.

\bibitem[\protect\citeauthoryear{Zhang \bgroup \em et al.\egroup }{2017}]{zhang2017regions}
Chao Zhang, Keyang Zhang, Quan Yuan, Haoruo Peng, Yu~Zheng, Tim Hanratty, Shaowen Wang, and Jiawei Han.
\newblock Regions, periods, activities: Uncovering urban dynamics via cross-modal representation learning.
\newblock In {\em Proceedings of the 26th International Conference on World Wide Web}, pages 361--370, 2017.

\bibitem[\protect\citeauthoryear{Zhang \bgroup \em et al.\egroup }{2021}]{zhang2021multi}
Mingyang Zhang, Tong Li, Yong Li, and Pan Hui.
\newblock Multi-view joint graph representation learning for urban region embedding.
\newblock In {\em Proceedings of the Twenty-Ninth International Conference on International Joint Conferences on Artificial Intelligence}, pages 4431--4437, 2021.

\bibitem[\protect\citeauthoryear{Zhang \bgroup \em et al.\egroup }{2024}]{zhang2024causally}
Yuyao Zhang, Ke~Guo, and Xiao Zhou.
\newblock Causally aware generative adversarial networks for light pollution control.
\newblock {\em arXiv preprint arXiv:2401.06453}, 2024.

\bibitem[\protect\citeauthoryear{Zheng \bgroup \em et al.\egroup }{2014}]{zheng2014urban}
Yu~Zheng, Licia Capra, Ouri Wolfson, and Hai Yang.
\newblock Urban computing: concepts, methodologies, and applications.
\newblock {\em ACM Transactions on Intelligent Systems and Technology (TIST)}, 5(3):1--55, 2014.

\bibitem[\protect\citeauthoryear{Zhou \bgroup \em et al.\egroup }{2017}]{zhou2017cultural}
Xiao Zhou, Desislava Hristova, Anastasios Noulas, Cecilia Mascolo, and Max Sklar.
\newblock Cultural investment and urban socio-economic development: a geosocial network approach.
\newblock {\em Royal Society open science}, 4(9):170413, 2017.

\bibitem[\protect\citeauthoryear{Zhou \bgroup \em et al.\egroup }{2018}]{zhou2018discovering}
Xiao Zhou, Anastasios Noulas, Cecilia Mascolo, and Zhongxiang Zhao.
\newblock Discovering latent patterns of urban cultural interactions in wechat for modern city planning.
\newblock In {\em Proceedings of the 24th ACM SIGKDD international conference on knowledge discovery \& data mining}, pages 1069--1078, 2018.

\bibitem[\protect\citeauthoryear{Zhou \bgroup \em et al.\egroup }{2019}]{zhou2019topic}
Xiao Zhou, Cecilia Mascolo, and Zhongxiang Zhao.
\newblock Topic-enhanced memory networks for personalised point-of-interest recommendation.
\newblock In {\em Proceedings of the 25th ACM SIGKDD International conference on knowledge discovery \& data mining}, pages 3018--3028, 2019.

\bibitem[\protect\citeauthoryear{Zhou \bgroup \em et al.\egroup }{2023}]{zhou2023phase}
Xiao Zhou, Xiaohu Zhang, Paolo Santi, and Carlo Ratti.
\newblock Phase-wise evaluation and optimization of non-pharmaceutical interventions to contain the covid-19 pandemic in the us.
\newblock {\em Frontiers in Public Health}, 11:1198973, 2023.

\end{thebibliography}

\end{document}